\documentclass[]{spie}  

\usepackage{amsmath,amsfonts,amssymb}
\usepackage{graphicx}
\usepackage[colorlinks=true, allcolors=blue]{hyperref}
\usepackage{cite}
\makeatletter

\title{Synthetic CT Skull Generation for Transcranial MR Imaging–Guided Focused Ultrasound Interventions with Conditional Adversarial Networks}

\author[a,$\dagger$]{Han Liu}
\author[b,c,$\dagger$]{Michelle K. Sigona}
\author[b,c,]{Thomas J. Manuel}
\author[c,d]{Li Min Chen}
\author[b,c,d]{Charles F. Caskey}
\author[e]{Benoit M. Dawant}
\affil[a]{Dept. of Computer Science, Vanderbilt University, Nashville, TN 37235, USA}
\affil[b]{Dept. of Biomedical Engineering, Vanderbilt University, Nashville, TN 37235, USA}
\affil[c]{Institute of Imaging Science, Vanderbilt University, Nashville, TN 37235, USA}
\affil[d]{Dept. of Radiology and Radiological Sciences, Vanderbilt University, Nashville, TN 37235, USA}
\affil[e]{Dept. of Electrical and Computer Engineering, Vanderbilt University, Nashville, TN 37235, USA}
\affil[$\dagger$]{These authors contributed equally to this work}

\begin{document} 
\maketitle

\begin{abstract}

\end{abstract}
Transcranial MRI-guided focused ultrasound (TcMRgFUS) is a therapeutic ultrasound method that focuses sound through the skull to a small region noninvasively under MRI guidance. It is clinically approved to thermally ablate regions of the thalamus and is being explored for other therapies, such as blood brain barrier opening and neuromodulation. To accurately target ultrasound through the skull, the transmitted waves must constructively interfere at the target region. However, heterogeneity of the sound speed, density, and ultrasound attenuation in different individuals’ skulls requires patient-specific estimates of these parameters for optimal treatment planning. CT imaging is currently the gold standard for estimating acoustic properties of an individual skull during clinical procedures, but CT imaging exposes patients to radiation and increases the overall number of imaging procedures required for therapy. A method to estimate acoustic parameters in the skull without the need for CT would be desirable. Here, we synthesized CT images from routinely acquired T1-weighted MRI by using a 3D patch-based conditional generative adversarial network (cGAN) and evaluated the performance of synthesized CT images for treatment planning with transcranial focused ultrasound. A dataset of 86 paired CT and T1-weighted MR images were randomly split so that 66 images were used for training, 10 for validation and parameter tuning, and 10 for acoustic testing. We compared the performance of synthetic CT (sCT) to real CT (rCT) images using an open-source treatment planning software, Kranion, and found that the number of active elements, skull density ratio, and skull thickness between rCT and sCT had Pearson’s Correlation Coefficients of 0.989, 0.915, and 0.941, respectively, suggesting strong positive linear correlation. Of a total of 990 elements $95.7\pm1.4\%$ of active and inactive elements overlapped between rCTs and sCTs. Simulations using the acoustic toolbox, k-Wave, resulted in $23.5\pm6.51\%$ less maximum root-mean-squared (RMS) pressure simulated with sCTs than the corresponding rCT pressure. An average focal shift of $0.96\pm 0.56$ mm and $1.07\pm 0.58$ mm was observed between the thalamus target and the maximum RMS pressure location in rCTs and sCTs, respectively. Our work demonstrates the feasibility of replacing real CT with the MR-synthesized CT for TcMRgFUS planning.


\section{INTRODUCTION}
\label{sec:intro}  
Transcranial MRI-guided focused ultrasound (TcMRgFUS) surgery is a novel noninvasive method of focusing energy through the skull that uses magnetic resonance imaging (MRI) for target identification, treatment planning, and closed-loop control of energy deposition\cite{1}. TcMRgFUS is clinically approved for unilateral thalamotomy\cite{ref1} and is being explored for other applications, such as drug delivery and neuromodulation\cite{ref2}.  Precise focusing is critical for TcMRgFUS to minimize treatment of off-target tissues\cite{2}. Prior to TcMRgFUS, CT images are acquired to estimate regional skull density, speed of sound and ultrasound attenuation during ultrasound wave propagation\cite{3}. CT imaging burdens patients by requiring longer screening time and increased risk due to radiation. Therefore, it is desirable to replace the real CT (rCT) images with synthetic CT (sCT) images that are generated from other imaging modalities. 

Deep-learning based methods have been previously used to generate synthetic CTs from MR images\cite{4}. Dual-echo ultrashort TE (UTE) MR imaging\cite{5} was used to train a 2D U-Net\cite{6} that was efficient at generating realistic skulls, but UTE scans are not always readily available. An alternative to UTE images are T1-weighted images, but these can be more challenging to synthesize CT skulls from than UTE because UTE imaging can capture signals from tissues with a very short transverse relaxation time such as bone, providing more information for skull synthesis. For instance, Lei \textit{et al.}\cite{7} proposed to use patch-based features extracted from MRIs to train a sequence of alternating random forests based on an iterative refinement model. Maspero \textit{et al.}\cite{8} trained three 2D cGANs\cite{cgan} for each plane and combine the results to generate synthetic CT from T1-weighted MRI. Gupta \textit{et al.}\cite{9} proposed to train a 2D U-Net on sagittal views of MRIs and synthesize the Hounsfield Unit (HU) of air, soft tissue and bone in three output channels. However, 2D networks are limited by the lack of information of relationship between slices\cite{2d} and the skulls are not spatially continuous (i.e generated volumes can appear jagged) along the views that are not involved in training. The irregular skull geometry of sCT may lead to significant differences in TcMRgFUS planning. In very recent work done in parallel to ours \cite{similar} a 3D cGAN was proposed to synthesize the whole head CT from MR images. Here we focus on the skull, which is the critical structure for TcMRgFUS.


Values from CT images of the head are used in different ways during treatment planning for TcMRgFUS. One important metric is the skull density ratio (SDR), an estimate of the transparency of the skull to ultrasound. The SDR is not always predictive of the energy needed to generate a focal spot transcranially, but a lower SDR is generally interpreted to mean lower acoustic transmission through the skull\cite{sdr,sdr2}. Although the precise method for computing SDR is proprietary, SDR is derived from the ratio of the HUs of trabecular to cortical bone along the line from a transducer element to the focus \cite{sdr2}. An open source software, Kranion, is available that is capable of generating SDR metrics highly correlated to those found in clinical procedures\cite{kranion}. Along with SDR, Kranion can report a skull thickness (ST) measurement between bone layers and number of elements (NAE) classified as 'active' (incident angle $<$ 20 degrees). For therapeutic applications, detailed computational models of human skulls have been created from CT and MR images to provide a map of acoustic parameters to model the propagation of sound through the skull using a full acoustic wave equation \cite{3,sun}. Using modeling tools like the acoustic toolbox, k-Wave\cite{kwave}, simulations are used to observe ultrasound waves interacting with subject-specific heterogeneous skulls, quantifying the focal shift, focus size, and energy loss caused by the aberrating skull. 

We hypothesize that synthetic CT generated from MRI can yield comparable clinical metrics for transcranial ultrasound that are derived from CT. Our study used two open-source software tools to compare skull metrics between sCT and rCT using 10 testing cases. We evaluated the performance of the rCT and sCT using Kranion to report the SDR, ST, and NAEs. Acoustic simulations were performed using k-Wave to calculate the pressure field formed from each CT. From each simulation we quantified the maximum intracranial pressure and focal shift from the target. Demonstration of similarity between sCT and rCT would show feasibility of synthesizing spatially continuous CT skulls from T1-weighted MRI. 

\begin{figure}[t]
\includegraphics[width=1\columnwidth]{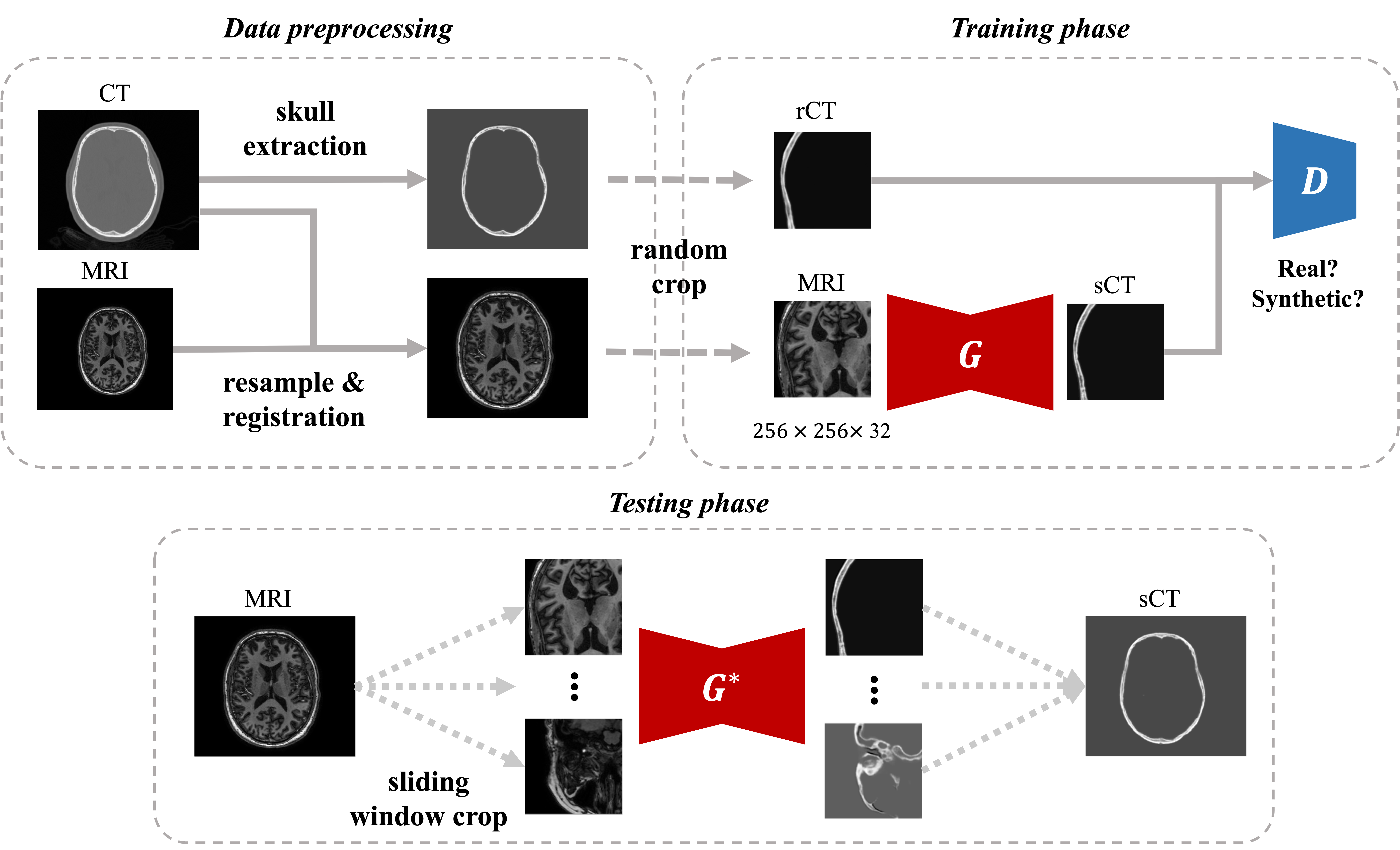}
\centering
\caption{An illustration of our proposed method.}
\label{fig1}
\end{figure} 

\section{METHODS}
Our dataset consisted of paired CT and T1-weighted MRI images of 86 human subjects. The in-plane resolution of CT images ranged from 0.4297 to 0.5449 mm and their slice thickness was 0.67 mm while the MR images had an isotropic voxel size of 1 mm. For data preprocessing, we first resampled and spatially aligned the low-resolution MR to the corresponding high-resolution CT by rigid registration. To extract the skull mask from the CT image, we used global thresholding with an intensity threshold of 400 HU and took the largest connected component. Then we applied morphological dilation to the binary skull mask to preserve contextual information for the skull. Lastly, the skull mask was used to filter the CT image to obtain the final CT skull image.

The overall workflow of our proposed method is illustrated in Figure 1. Our network was a 3D patch-based cGAN which enabled paired training to synthesize CT skull images from the T1-weighted MRI. For our generator $G$, we used a 3D ResNet\cite{res} with 9 residual blocks. In the output layer of $G$, we applied a Tanh activation which allowed the model to learn more quickly to saturate\cite{dcgan} and multiplied the output by 2000 to bound the range of the sCT intensity to $[-2000, 2000]$ HU.  For our discriminator $D$, we used a 3D convolutional PatchGAN classifier\cite{cgan}. The network input was a single-channel 3D patch $x$ with the dimension of $256 \times 256 \times 32$ voxels that is randomly cropped from the original MRI volume $X$. The corresponding CT patch $y$ was cropped from the same spatial position from the original CT volume $Y$. During training, $G$ aimed to generate synthetic CT skull image $G(x)$ from $x$ to fool $D$, while $D$ learned to classify $y$ and $G(x)$. Our loss function was composed of an adversarial loss $L_{cGAN}$ and a L1 loss $L_{L1}$ which encouraged less blurring (in Koh \textit{et al.}\cite{similar}, an additional gradient loss was incorporated to generate sCT for the whole head):

\begin{equation}
    L_{cGAN}(G,D)=\mathbb{E}_{x,y}[log(D(x,y))]+\mathbb{E}_{x,z}[log(1-D(x,G(x,z)))]
\end{equation}

\begin{equation}
    L_{L1}(G)=\mathbb{E}_{x,y,z}[||y-G(x,z)||_{1}]
\end{equation}

We used the Adam optimizer to update the weights of $D$ and $G$ alternatively. We set an initial learning rate as $2\times 10^{-4}$ which decayed to zero over 100 epochs. A minibatch size of 1 was used. The intensities of MRI volumes were rescaled  to $[0, 1]$ before being cropped into patches. Random intensity shifts between $-10$ and $10$ percent and random gamma adjustment with gamma in the range of $0.5$ to $1.5$ were used for data augmentation. The augmentation strategy encouraged the network to learn to map a dynamic MR intensity distribution to a fixed CT HU distribution. During the inference phase, since our network was trained with randomly cropped patches, the whole synthetic CT was generated using a sliding window. The strides for the adjacent patches are 32, 32 and 8 voxels.

In our experiment, we randomly split the entire dataset into subgroups containing 66 images for training, 10 for validation and the remaining 10 for testing. To avoid overfitting, we used the validation set to tune the hyperparameters and selected the best sCT generator $G^{*}$. We selected $G^{*}$ from the epoch in which the Mean Squared Error (MSE) between the rCT and sCT were the smallest in the validation set. To assess the quality of the sCT, we evaluated the image similarity between sCT and rCT by calculating the Mean Absolute Error (MAE) in HU. 

In Kranion, we placed a virtual 990-element hemispherical array transducer so that the geometric focus was
at either the left or right thalamus. This transducer modeled the 1024-element ExAblate transducer (Insightec, Tirat Carmel, Israel) with 34 elements deactivated or containing passive cavitation dectectors. The transducer was tilted along the x and y axes to maximize the number of active elements with the sCT without exceeding 10 degrees in each direction to maintain a realistic scenario. We targeted the left and right thalamus in rCT and sCT separately so that there were two targets for each skull in our test data set of 10 skulls. For each virtual targeting procedure (N=20), we measured the NAE, the SDR, and the ST (the distance between skull boundaries along a ray path) between the rCT and sCT. 

The CT, MR, transducer element positions, and focus position were exported from Kranion (Figure 5A) and imported to MATLAB to run full-wave acoustic simulations using the open-source toolbox k-Wave\cite{kwave}. The skull was incorporated in simulations using a linear approximation to map HU to bone porosity, and then calculated speed of sound, density and attenuation from porosity  ($c_{max} = 3100 m\cdot s^{-2}, c_{min} = 1480 m\cdot s^{-2}, \rho_{max} = 2100 kg\cdot m^{-3}, \rho_{min} = 1000 kg\cdot m^{-3}, a_{0} = 8.1 dB \cdot cm^{-1} \cdot MHz^{-b}, b=1.1$)\cite{3}. Images were padded to match a simulation grid of [625, 625, 405] with isotropic grid spacing of 0.5 mm. Simulations were performed at 650 kHz, maintaining a spatial discretization greater than 7 points per wavelength in water. To minimize simulation time, a 100 cycle waveform was used. All simulations were run on a Quadro P6000 GPU (NVIDIA Corporation, Santa Clara, CA). The root-mean-squared (RMS) pressure was recorded for each voxel location. To evaluate the simulated results, the maximum intracranial RMS pressure and focal shift were calculated. The left or right thalamus target was compared with the location of the maximum pressure for each simulation to calculate the focal shift. 

\begin{figure}[h]
\includegraphics[width=1\columnwidth]{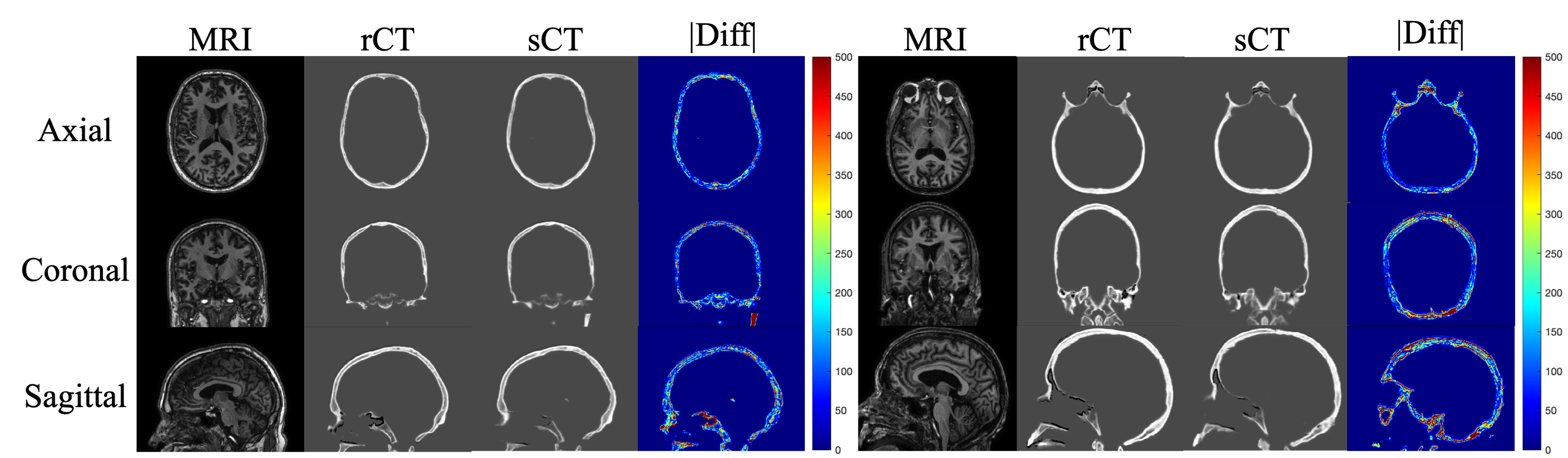}
\centering
\caption{Examples of the original T1-weighted MRI, real CT (rCT), synthetic CT (sCT) and absolute difference map ($|$diff$|$). The two subjects with MAE of 157.98 HU (left) and 233.76 HU (right) in the skull regions are shown. For visualization purpose, we rescaled the difference heatmap to $[0, 500]$ HU. Notice that the bones in the inferior brain area tend to have a larger difference.}
\label{fig2}
\end{figure} 

\section{RESULTS}
In our testing set, the MAE between rCTs and sCTs in skull regions was $190.94\pm22.40$ HU. Absolute difference heatmaps for test cases with the smallest and largest MAE exhibited larger MAE at the inferior part of the skull than the superior part (Figure 2). Note that we used 3D networks and thus our synthesized skulls were spatially continuous in all views. 

\begin{figure}[t]
\includegraphics[width=1\columnwidth]{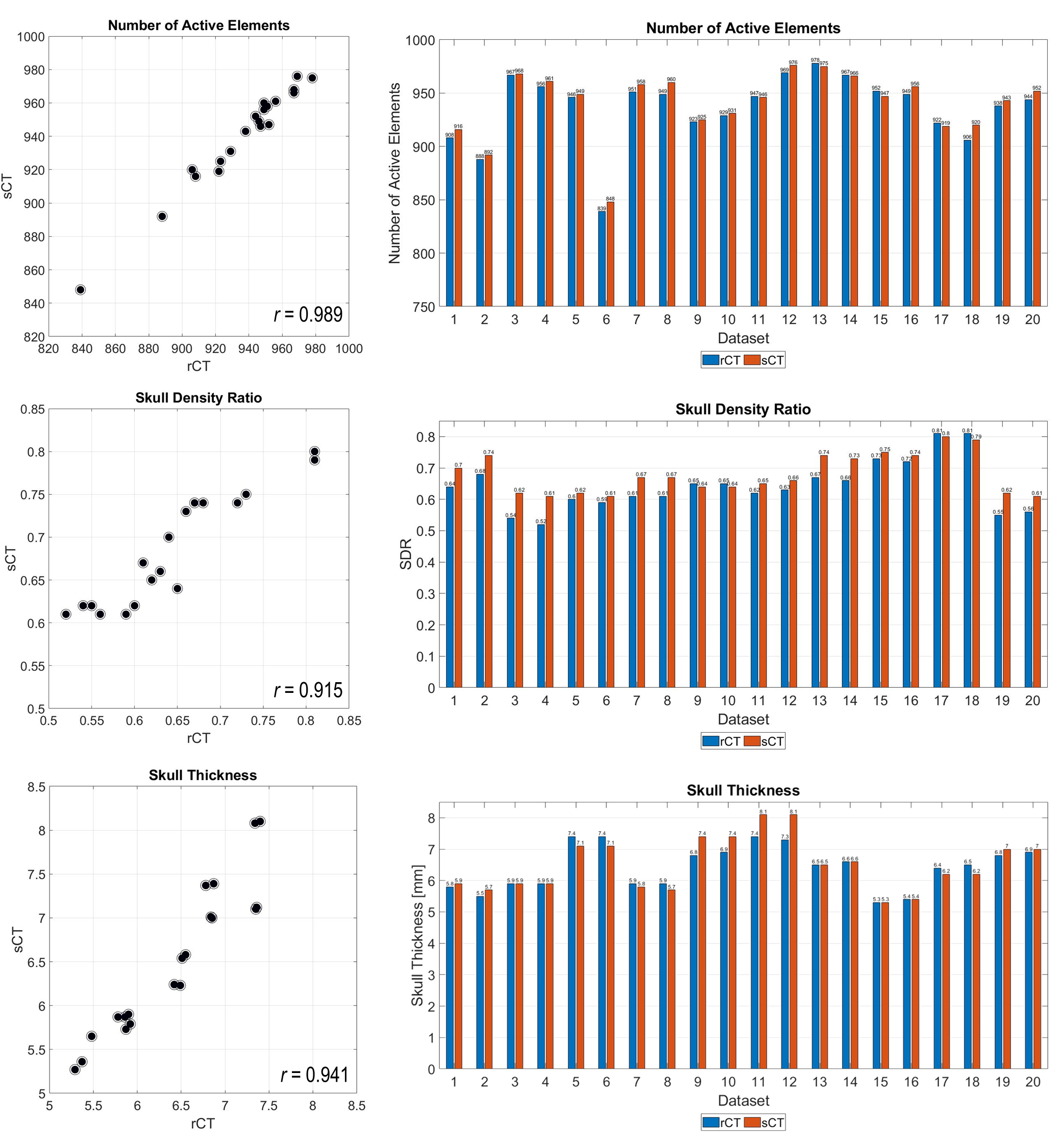}
\centering
\caption{The scatter plots and bar charts of the number of elements, skull density ratios, and skull thickness between rCT and sCT for 20 testing simulation targets.}
\label{fig3}
\end{figure} 

\begin{figure}[h]
\includegraphics[width=1\columnwidth]{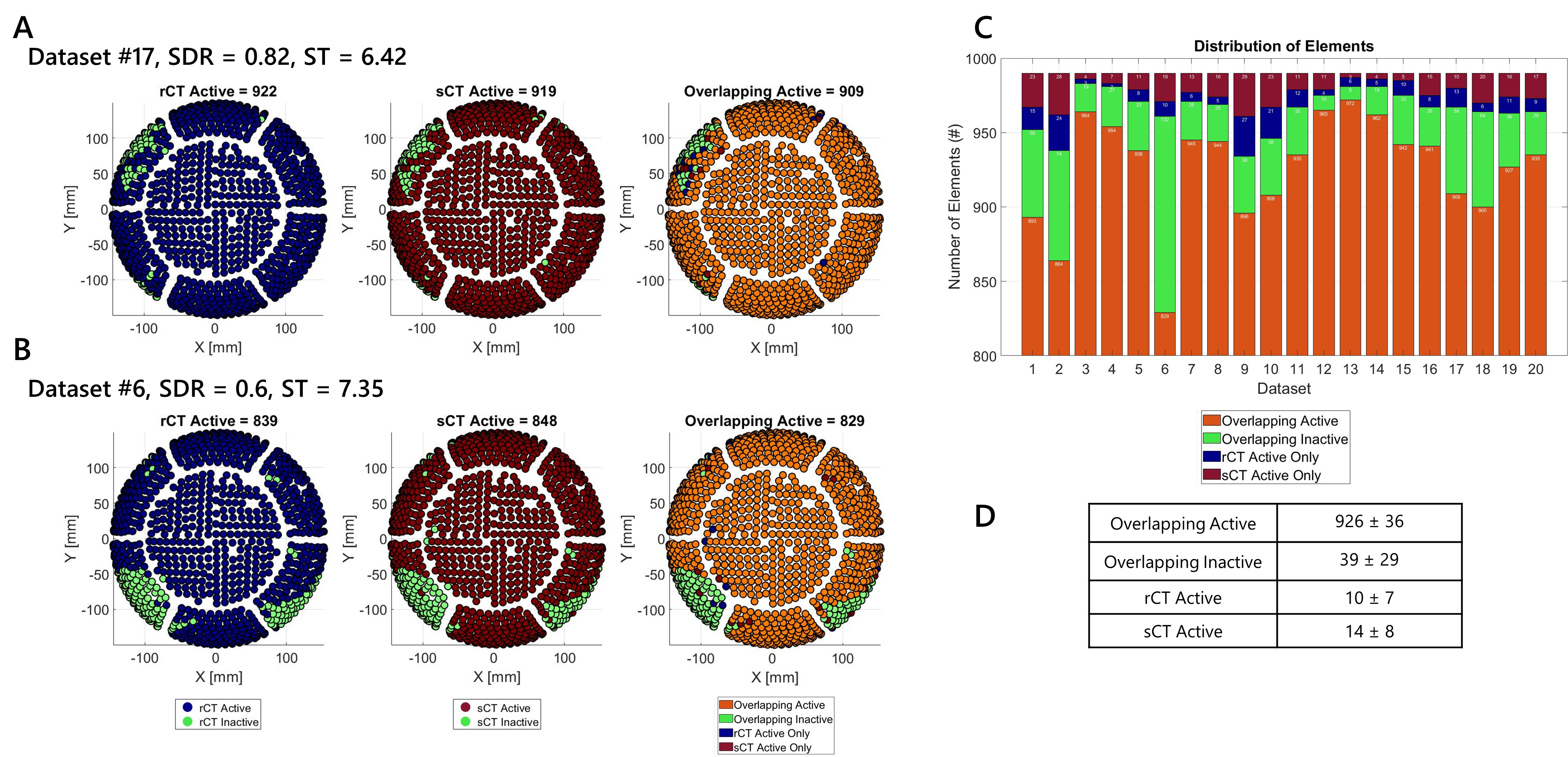}
\centering
\caption{Color coded plots showing the distribution of active and inactive elements of the hemispherical array for a high SDR group (A) and low SDR group (B).  From left to right, the active and inactive elements determined by Kranion are calculated with rCT, sCT, and the overlapping and non-overlapping elements between rCT and sCT. The number of active elements are reported in the title of each plot. The distribution of elements for all cases are shown in the stacked bar graph (C), corresponding with the same colors from A and B. The $mean\pm STD$ for all cases were calculated for each subgroup and shown in the table (D). }
\label{fig4}
\end{figure} 

\begin{figure}[h]
\includegraphics[width=1\columnwidth]{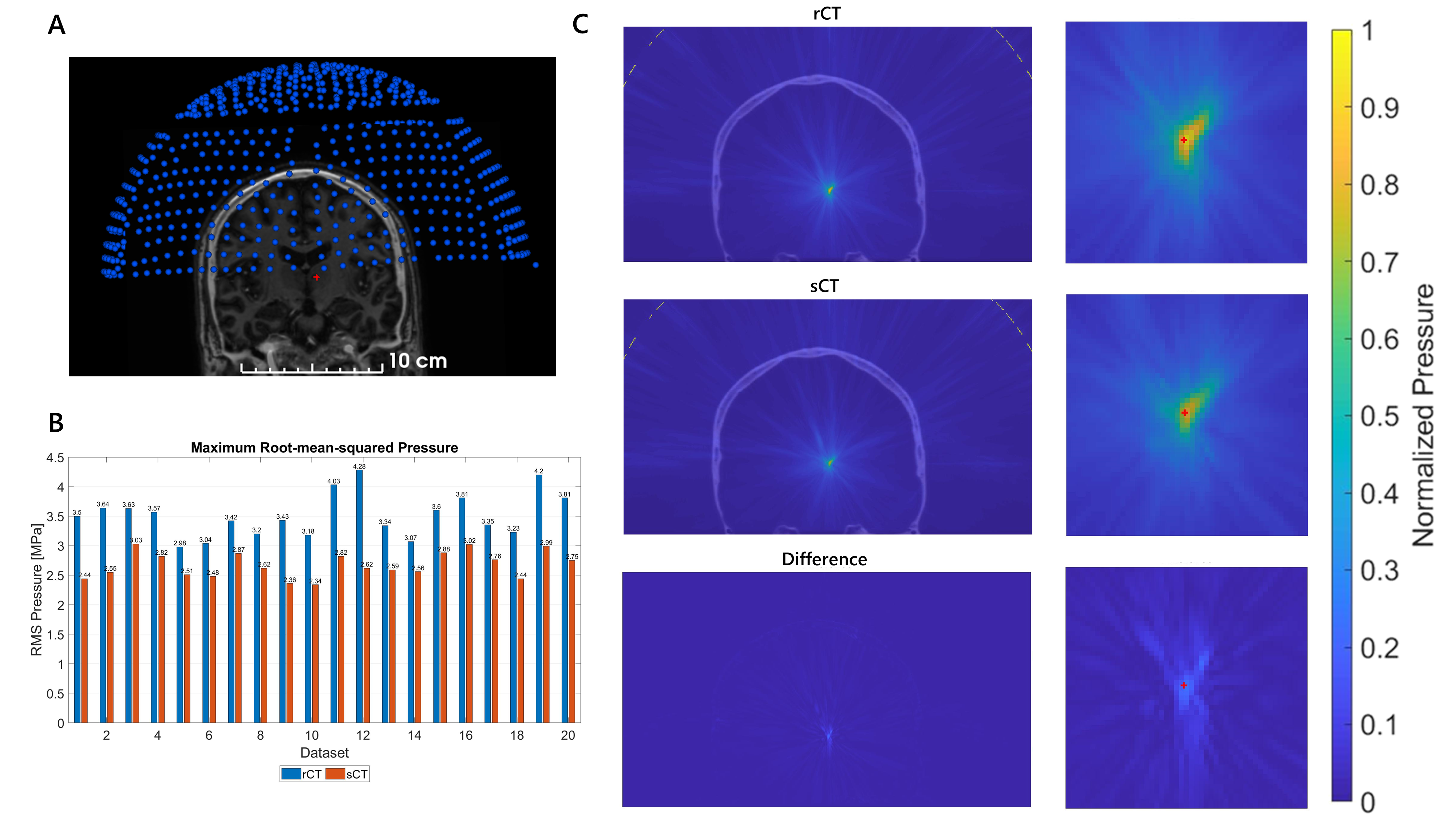}
\centering
\caption{MR, sCT, transducer element positions (blue spheres), and focus position (red cross) exported from Kranion (A). When exported, all files are registered to a local coordinate system which integrates with k-Wave for simulations. A bar plot of all 20 cases is shown (B). It can be noted that the sCT maximum RMS pressure is lower across all cases. Orthogonal XZ planes of acoustic simulations from case \#3 (SDR = 0.54, ST = 5.9) are shown normalized to the maximum RMS pressure from the  rCT simulation (C). The absolute difference between rCT and sCT are shown on the bottom, and planes zoomed in on the focus are shown on the right column. The intended target is indicated by the red cross.}
\label{fig5}
\end{figure} 

Metrics from Kranion exhibit strong similarity between rCT and sCT across all cases (Figure 3). The Pearson’s Correlation Coefficients for the NAE, SDR, and ST are 0.989, 0.915, and 0.941 respectively, demonstrating strong positive linear correlation of these metrics between rCT and sCT. Among the 20 simulated targets, the differences of NAE, SDR , and ST are $5.30\pm3.56$, $0.04\pm0.03$, and $0.22 \pm 0.23$, equal to nearly $0.6\%$, $6.7\%$, and $3.5\%$ of the rCT NAE, SDR, and ST. Further evaluation of NAEs revealed the majority of active and inactive elements ($97.5\pm1.4\%$) overlapped between rCT and sCT images (Figure 4A and 4B). Of a total of 990 elements, $926 \pm 36$ active elements and $39 \pm 29$ inactive elements were overlapping (Figure 4C and 4D). The pressure fields in simulations of the rCT versus sCT contained a lower RMS pressure at the maximum focus ($23.5\pm6.5\%$ ) (Figure 5B) and minimal differences between rCT and sCT observed in the orthogonal slices (Figure 5C). Comparing the maximum RMS pressure location with the thalamus target, the distance vector was approximately 1 mm for both skull sets ($0.97 \pm 0.56$ mm between rCT and  target and $1.07\pm 0.58$ mm between sCT and target), a mean difference of $0.35\pm 0.40$ mm between the rCT and sCT maximum RMS pressure locations. 

\section{DISCUSSION and CONCLUSIONS}
In this work, we explored the feasibility of replacing real CT with synthetic CT images for TcMRgFUS planning. We showed that the synthesized CT skull generated by our proposed 3D patch-based cGAN are (1) spatially continuous and (2) highly comparable to real CTs when used with a TcMRgFUS simulator. Through Kranion, skull metric and element comparisons were shown to have strong correlation between rCT and sCT. Acoustic simulations revealed slight discrepancies between rCT and sCT in maximum pressure at the focus and focal location. 

Similar work using a 3D cGAN from Koh et al.\cite{similar} observed a larger MAE than ours ($190\pm22$ vs. $280\pm24$) but a larger correlation coefficient between the real and synthesized SDRs (0.92 vs. 0.95). It was observed MAE was greater at the inferior of the skull than the superior of the skull. We conjecture that the inferior has a greater MAE because the anatomical structures at the inferior part of the head are more complex than those at the superior part and thus make our image synthesis task more difficult. 

Characterizing the sCTs using acoustic simulations provided further information about the clinical potential of sCTs. Koh et al. showed smaller percent error of maximum pressure at the focus ($23.5 \pm 6.51\%$ vs. $3.11\pm 2.79\%$). Some important differences between our study and Koh et al. are (1) our simulations were performed at a higher frequency (650 kHz vs. 200 kHz) (2) we used a transducer modeling the ExAblate transducer rather than a single-element transducer (990 elements vs. 1 element). Using a higher fundamental frequency is beneficial because a smaller wavelength results in a focal size with finer spatial resolution. Because attenuation is frequency dependent, at higher frequencies we would expect a greater decrease in intensity at the focus. The hemispherical array encompasses about half of the skull (Figure 5A) and because all elements were used in these acoustic simulations, the interactions with the skull would provide greater contributions to the pressure field rather than a single-element transducer with a smaller surface area. Regarding the lower maximum pressure observed in our study for sCTs than rCTs, a sensitivity analysis studied the intracranial peak pressure changes with varying smoothing kernels as observed from clinical CT scanners and found a smoothing kernel of 0.35-0.8 mm could cause significant errors of peak pressure of $5\%$, increasing with higher frequencies\cite{sensitivity}. This suggests that the rCTs may have less variation within the skull than sCTs, reducing reflections at the interface. 

A few limitations of acoustic simulations were identified in our study. To minimize computational time, skulls in all simulations were cropped so that the base of the skull below the target was not included in the simulation grid. This limits the contributions to the pressure field from reflections off the skull base. It can be noted that a simulation study\cite{mueller} showed that inclusion of the skull base is less important for large skulls, but more important for smaller skulls like nonhuman primates or rodents. Multi-element arrays are appealing for TcMRgFUS procedures because phase correction can be performed. Multi-element arrays allow independent phase-correction of individual elements to correct for large acoustic impedance mismatches between the skull-tissue boundaries, restoring energy at the target\cite{phase}. Phase correction was not applied and evaluated in the context of our study, but can be investigated in the future with Kranion \cite{kranion2}. This work did not include thermal simulations to evaluate temperature heating at the skull surface and in the brain, but should be studied in the context of ablation treatments and non-thermal treatments like neuromodulation. This study is ongoing as well as a study in which clinical focused ultrasound plans used for lesioning are compared to plans produced with synthesized CTs. While we have drawn comparisons between our study and Koh's work, a comparative study would be necessary to decide if observations in MAE and acoustic simulation results are because of differences in approach or differences in data sets. 

Our work evaluated CT images synthetically generated from 3D cGAN with real CT scans for potential use in TcMRgFUS procedures. CTs are currently the gold standard to provide subject-specific skull properties that are necessary for treatment planning of TcMRgFUS. Our study showed that sCTs generated from the  3D cGAN network could replace the need for CT scans with a routine T1-weighted MR image. Patient selection for TcMRgFUS is assessed by metrics characterizing the skull, which we showed sCTs are comparable to rCTs for all skulls. Acoustic simulations allow for subject-specific estimation of the pressure field emitted by the transducer and are necessary to assess targeting and calculate phase correction in TcMRgFUS procedures. Replacement of sCTs with rCTs would decrease patient burden of additional scan time and minimize exposure to radiation.

\section{ACKNOWLEDGEMENTS}
This work has been supported by NIH U18EB029351 and the Advanced Computing Center for Research and Education (ACCRE) of Vanderbilt University. All acoustic simulations were run on a Quadro P6000 GPU donated by NVIDIA Corporation. The content is solely the responsibility of the authors and does not necessarily represent the official views of these institutes.

\end{document}